%% file: main.tex
\pdfoutput=1

\documentclass[11pt]{article}

\usepackage[final]{acl}
\usepackage{float}
\usepackage{CJKutf8}
\usepackage{makecell}
\usepackage{times}
\usepackage{latexsym}
\usepackage{tabularx}
\usepackage{xcolor}
\usepackage{subcaption}
\usepackage{hyperref}
\usepackage{amsmath}

\usepackage[T1]{fontenc}

\usepackage[utf8]{inputenc}

\usepackage[inkscapelatex=false]{svg}

\usepackage{microtype}
\usepackage{enumitem}
\usepackage{inconsolata}
\usepackage{booktabs} 

\usepackage{graphicx}

\title{Do LLMs Understand Wine Descriptors Across Cultures? \\
A Benchmark for Cultural Adaptations of Wine Reviews}

\author{
 \textbf{Chenye Zou\textsuperscript{1}},
 \textbf{Xingyue Wen\textsuperscript{2}},
 \textbf{Tianyi Hu\textsuperscript{1,4}},
 \textbf{Qian Janice Wang\textsuperscript{3}},
\\
 \textbf{Daniel Hershcovich\textsuperscript{1}}
\\
\\
 \textsuperscript{1}Department of Computer Science, University of Copenhagen \\
 \textsuperscript{2}Department of Food and Resource Economics, University of Copenhagen
\\
 \textsuperscript{3}Department of Food Science, University of Copenhagen \\
 \textsuperscript{4}Department of Computer Science, Aarhus University
\\
 \small{{zoucy2001@gmail.com}}
}

\begin{document}
\maketitle
\begin{abstract}
Recent advances in large language models (LLMs) have opened the door to culture-aware language tasks. We introduce the novel problem of adapting wine reviews across Chinese and English, which goes beyond literal translation by incorporating regional taste preferences and culture-specific flavor descriptors. In a case study on cross-cultural wine review adaptation, we compile the \textbf{first} parallel corpus of professional reviews, containing 8k Chinese and 16k Anglophone reviews. We benchmark both neural-machine-translation baselines and state-of-the-art LLMs with automatic metrics and human evaluation. For the latter, we propose three culture-oriented criteria—Cultural Proximity, Cultural Neutrality, and Cultural Genuineness—to assess how naturally a translated review resonates with target-culture readers. Our analysis shows that current models struggle to capture cultural nuances, especially in translating wine descriptions across different cultures. This highlights the challenges and limitations of translation models in handling cultural content.
\end{abstract}

\input{latex/Introduction}

\input{latex/Related_works}

\input{latex/Datasets}

\input{latex/Experiment}

\input{latex/Results}

\input{latex/Conclusion}

\input{latex/Limitation}

\section*{Ethical Considerations}
This study relies on wine ratings and tasting notes sourced from professional websites. All datasets are publicly accessible and do not involve any user privacy. Our experiments strictly comply with the terms of use of these platforms.

To support transparency and reproducibility, we release all source code and experimental configurations. In addition, we publicly share the portion of the dataset that does not contain proprietary content, accessible through open-access sources and subject to applicable licensing terms.
We license our data under the CC0 1.0 DEED and require users to sign an agreement restricting its use to academic research, further protecting potential user interests.



\section*{Acknowledgments}
Thanks to the anonymous reviewers and action editors for their helpful feedback. The authors express their gratitude to Haixin Qi for his assistance with the dataset processing.

\bibliography{main}

\appendix

\input{latex/Appendix}

\end{document}

%% file: latex/Introduction.tex
\section{Introduction}
\begin{figure}[t]
    \centering
    \includegraphics[width=\linewidth]{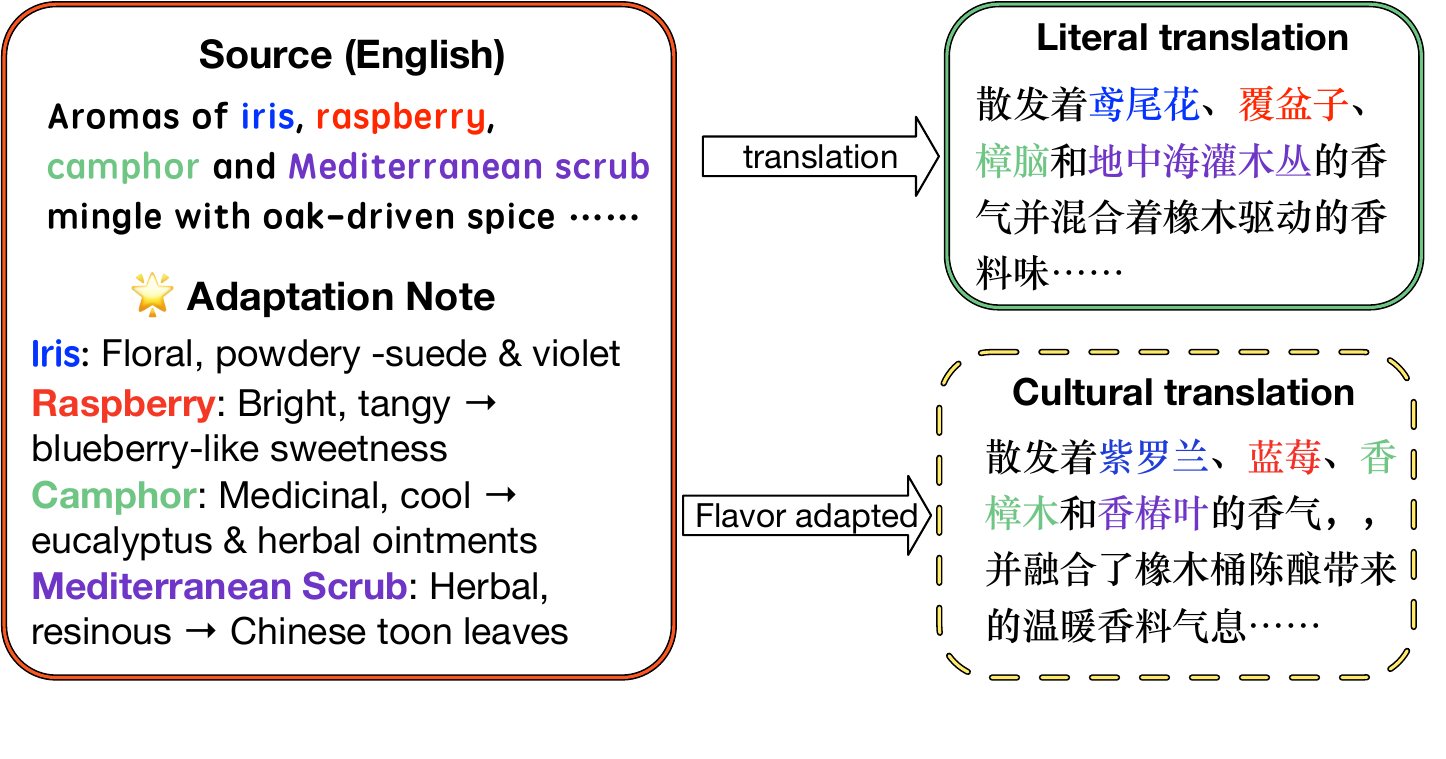}
    \caption{Culture-specific flavor misinterpretation in translation. Color coding highlights shifts in meaning: {\color{blue}blue} represents floral, {\color{red}red} represents fruit, {\color{green}green} represents spice, and {\color{violet}purple} represents vegetal reinterpretations. The figure shows how culturally unfamiliar terms may be recontextualized in target-language adaptations.} 
    \label{fig:example} 
\end{figure}
Wine reviews serve as valuable guides for consumers, offering detailed insights into the characteristics of each bottle. For casual drinkers and connoisseurs alike, these reviews act as a compass, helping them navigate the vast selection of wines available. However, due to cultural influences and geographical distinctions, beverage consumption patterns and individual preferences vary significantly across regions \citep{rodrigues2019contribution}, not to mention reviews. These variations extend beyond mere differences in taste and are deeply rooted in the cultural, social, and environmental contexts of each region. Consequently, consumer preferences for certain beverages, including wine, can differ greatly, often rendering generalized reviews less relevant or applicable across diverse audiences. Professional reviews play a greater role than user reviews in promoting consumer purchases \citep{chiou2014whose}. For Chinese wine consumers, professional wine reviews in Chinese are scarce, and most available reviews require a paid subscription. Similarly, Western consumers face challenges finding professional reviews of Chinese wines. Yet, interest in Chinese wine is rapidly growing \citep{wineintelligence2024,decanter2023}. Consumer-research evidence underscores why accessible, comprehensible reviews matter \citep{danner2017like, spence2020wine}. These trends underscore the increasing global relevance of Chinese wine and highlight the need for better information accessibility and cross-cultural understanding.  While some may regard wine tasting as a niche domain, recent works have shown that food and drink constitute an active and important area of NLP research \citep{adilazuarda-etal-2024-towards,li-etal-2024-foodieqa,survey,zhou-etal-2025-mapo}, as they provide rich, culturally nuanced language ideal for studying subjectivity and cultural adaptation. Despite this growing interest, existing studies have not addressed the specific challenges of cross-cultural adaptation in professional wine tasting notes, which are prototypical examples of culturally shaped subjective expressions. Our work fills this gap by providing a specialized dataset and empirical analyses, offering insights that extend beyond wine to other food, drink, and culturally sensitive domains in NLP.

Our initial goal is to address the following research question \textbf{RQ1:} To what extent do Chinese and Western professional wine reviews differ? To answer \textbf{RQ1}, we propose the first cross-cultural multilingual wine reviews dataset  CulturalWR, covering over 5k wines and about 25k reviews. We then perform comprehensive data analysis on it to analyze lexical and semantic differences. We demonstrate that Chinese and Western reviewers tend to emphasize different flavor descriptors and adopt divergent stylistic conventions in professional writing. These findings highlight the need for cultural adaptation when translating wine reviews across languages. This paper aims to bridge these gaps by providing a comprehensive, culturally inclusive dataset of bilingual wine reviews, supporting a more globally relevant perspective on wine preferences.

Recognizing and adapting to cultural differences in language use is both essential and challenging \citep{hershcovich-etal-2022-challenges}, especially for subjective comments. Translations of reviews using current neural machine translation systems may overlook culture-specific expressions or result in mistranslations due to insufficient grounding in physical and cultural contexts. For instance, in Figure \ref{fig:example}, `raspberry', a common European flavor descriptor, is rare in China (`\begin{CJK}{UTF8}{gbsn}覆盆子\end{CJK}'), making its taste unfamiliar to Chinese consumers. The flavor of raspberry is puzzling to Chinese consumers, and requires additional explanation or finding a fruit with a similar flavor. `Blueberry', which has a similar flavor profile  \citep{jin2022chinese}, is more popular and better understood by Chinese consumers. All the outputs of this example are shown in Appendix \ref{sec:examples}. 

Based on the findings, we are interested in answering the research question \textbf{RQ2:}, To what extent do cross-cultural flavor-descriptor differences hinder target-culture consumers’ comprehension of wine reviews? To measure this, we introduce three culture-related evaluation metrics - Cultural Proximity, Cultural Neutrality, and Cultural Genuineness and conduct a human assessment to analyze different cultures' reviews from these perspectives. We found that readers most easily understand reviews rooted in their own culture. Surprisingly, Western reviews are harder for Chinese readers, whereas Chinese reviews are rated slightly easier to understand by Western readers.

Finally, given the lack of sensory grounding in LLMs \cite{abdou-etal-2021-language,patel2022mapping}, we are interested in answering the research question \textbf{RQ3:} To what extent do LLM-generated culturally adapted wine reviews enhance cross-cultural alignment and reader comprehension compared to human translations? To answer this question, we prompted the LLM with culture-aware instructions to translate the reviews and then conducted a blind human evaluation. According to the annotators' ratings, the LLM can, under appropriate prompting, sometimes outperform literal human translations—a surprising outcome.

In this work, we introduce the task of adapting wine reviews across languages and cultures. Beyond direct translation, this requires adaptation concerning content and style. We focus on Chinese and English wine reviews, automatically pairing reviews for the same wine from the same vintage from two monolingual corpora. As there are many reviews in English for the same wine, we also explore the inner differences in reviews from people of the same culture. We evaluate our methodology with human evaluation and automatic evaluations on the dataset we construct. Our contributions are as follows:
\begin{enumerate}
    \item We introduce the novel task of cross-cultural wine review translation and build CulturalWR, the \textbf{first} bidirectional Chinese-English dataset with multiple references. 
    
    \item We conduct a detailed evaluation and analysis of the cultural differences between Chinese and English-speaking communities in the lexical choices, semantic differences, and sensory descriptions used in wine reviews.

    \item We experiment with various sequence-to-sequence approaches to adapt the reviews, including machine translation models and multilingual LLMs.\footnote{Data and code available in \href{https://github.com/IAN-YE/CultureWR.git}{https://github.com/IAN-YE/CultureWR.git}}
\end{enumerate}


%% file: latex/Related_works.tex
\section{Related Work}
\paragraph{Computational analysis of wine reviews.}
Recent advancements in wine informatics have leveraged computational techniques to analyze expert wine reviews. The Computational Wine Wheel 2.0 facilitates machine-learning-based wine-attribute analysis \citep{chen2016computational}. Machine-learning and text-mining techniques have been utilized to discover predictive patterns in wine descriptions, challenging the notion that flavor descriptions are purely subjective \citep{lefever-etal-2018-discovering}. Recent work has begun to interrogate linguistic and cultural variation in wine discourse. \citep{horberg2025chemo} used a 68-million-word corpus and transformer language models to map chemosensory vocabularies across wine, perfume and food reviews, revealing domain-specific semantic clusters. Cross-lingual resources have also emerged: Wang et al. (2021) released an English–Chinese parallel corpus of 1,211 aligned reviews, enabling systematic comparison of culture-specific tasting terms \citep{wang2021parallel}. Sensory studies confirm that cultural familiarity shapes perceived quality, as shown in a comparative experiment with British and Spanish experts \citep{suarez2023culture}. These studies provide a foundation for computational analysis of wine reviews, yet they primarily focus on prediction tasks rather than cross-cultural aspects of wine appreciation.

\paragraph{Cross-cultural analysis of flavors.} Flavor perception varies across cultures, as shown in studies analyzing beer pairing preferences in Latin America \citep{ARELLANOCOVARRUBIAS2019303} and color-flavor associations in snack packaging across China, Colombia, and the UK \citep{VELASCO201449}. Research on cheddar cheese flavor lexicons across different countries \citep{DRAKE2005473} further highlights cultural influences on taste perception. consumer conceptualisations of “red” and “white” diverge sharply between French, Portuguese and South-African drinkers \citep{fairbairn2024concept}, while translation studies demonstrate how Western terroir terms acquire auspicious connotations in Chinese branding \citep{niu2023terroir}. These findings underscore the need for culturally adaptive approaches in wine description and translation.
\paragraph{Cultural adaptation.} As culture and language are intertwined and inseparable, there is a rising demand to equip machine translation systems with greater cultural awareness \citep{NITTA1986101,ostler-1999-limits,hershcovich-etal-2022-challenges}. However, it is costly and time-consuming to collect culturally sensitive data and perform a human-centered evaluation \cite{liebling-etal-2022-opportunities}.
Cultural adaptation in NLP  focuses on adjusting text style while preserving its original meaning \cite{coli_a_00426}. This is particularly important in cross-cultural tasks, where differences in rhetorical structures and communicative conventions influence translation. For culturally-aware translation, eliciting explanations has been shown to significantly enhance comprehension of culturally specific entities \citep{yao2024benchmarkingmachinetranslationcultural, Singh_2024}.
Recent studies showed that LLMs are adept at adapting cooking recipes across cultures, using a comparable and a parallel corpus \citep{cao2023culturaladaptationrecipes,hu2024bridging}. Our work focuses on the translation of non-parallel subjective reviews, an area that remains underexplored. 

%% file: latex/Datasets.tex
\section{Data Collection \& Analysis}
We present CulturalWR, a dataset of paired professional wine reviews. Here we describe how we construct our dataset. Each pair consists of one Chinese review by Chinese wine critics and, when available, an English review by the same authors, alongside multiple English reviews authored by Western wine critics for the same wine, identified by the wine name and its corresponding vintage.
\subsection{Data Collection}
We collect the English wine reviews from one wine sales website, Wine, and a vertical search engine, Wine-Searcher. Chinese wine reviews are primarily written by two prominent reviewers, Alexandre Ma\footnote{\url{https://www.alexandrema.com}}, Shen Hao\footnote{\url{http://www.leparadisduvin.com}}, who often publish bilingual wine reviews and China Wine Information Network\footnote{\url{www.winesinfo.com}} comprising 35 Chinese wine experts.

\subsubsection{Data preprocessing}
Wine names are usually derived from regions or grape varieties, labeled by these features or a unique name. We observed that Chinese reviewers often skip mentioning the grape variety and region, opting for simpler names. We use string matching with character normalization (e.g., converting "Château Nénin" to "Chateau Nenin") to match reviews to specific wines. 
Exact matches are assumed to indicate the same wine; partial matches, which total 1,027, undergo manual verification.\footnote{Particularly in the Bordeaux region, wines use the winery's name for the Grand Vin (first wine) and unique names for second and third wines, with "blanc' added for white wines.} In our manual verification of 1,027 partial matches, we confirmed 838 instances as correct matches and rejected 189 instances due to discrepancies in wine type, vintage year, or ambiguity arising from similar winery names used across different labels (e.g., Banfi Centine vs. Banfi Centine Bianco are different wines, whereas Blason d’Issan and Château d’Issan Blason d’Issan refer to the same wine). After confirming the names and vintage years match, we finalize the reviews for each specific wine. 

To focus on red wines and adapt reviews culturally, we applied three filtering rules: excluding Non-Vintage (N.V.) wines due to inconsistent tasting notes over time, separating white and rosé wines into distinct datasets for independent testing, and filtering out reviews under 30 words in English or 30 characters in Chinese for lack of detail.

\subsection{Dataset Overview} We collect approximately 20k Chinese wine reviews covering nearly 20k wines, along with 150,000 English wine reviews spanning over 30k wines. These reviews encompass a variety of wine types, including red, white, rosé, and champagne. After filtering, we retain around 10k Chinese reviews focused on red wines. Through matching, the dataset is refined to include 4.5k wines, comprising about 4.5k Chinese reviews and 16k English reviews. 3,227 Chinese reviews have their matched English reviews written by the same author. Data statistics are shown in Table \ref{tab:review_stats}. 

Besides English reviews, we also collected some reviews written in German, Portuguese, French, Dutch, Italian and Spanish. And the statistics of the reviews reported in Appendix \ref{sec:Dataset}. we use langdetect\footnote{\url{https://pypi.org/project/langdetect/}} to identify these languages.

\begin{table}[h]
\centering
\resizebox{\linewidth}{!}{ 
\begin{tabular}{c|c|c}
\toprule
                                & Number  & Mean \#Tokens \\ \hline
CA Chinese Reviews & 4776    & 67.57       \\     
CA English Reviews & 3227    & 74.25       \\ \midrule
Transl. Chinese Reviews & 60 & 60.2    \\ \midrule
WA English Reviews & 16746  & 58.16       \\ \bottomrule

\end{tabular}
}
\caption{Statistics of reviews. CA refers to Chinese wine critics and WA to Western ones. We count tokens with \texttt{jieba} text segmentation for Chinese and whitespace tokenization for English.}
\label{tab:review_stats}
\end{table}

\textbf{Attributes} Besides the basic reviews and their corresponding rating for each wine, we sourced wine data from different sources, we reorganized the basic attributes of all these wines, which includes the geographical location of the winery, the composition of the grape varietals, the vintage, the alcohol content and the price. To ensure privacy, we anonymized the obtained data, and no personally identifiable information is included in the attributes. See Appendix \ref{sec:attributes}.

%% file: latex/Experiment.tex
\section{Cross-Cultural Wine Reviews Adaptation Task}
We propose cross-cultural wine review adaptation, extending machine translation by requiring both accuracy and deliberate semantic divergence to address cultural differences.

Evaluating cultural adaptation is challenging, as it must balance meaning preservation with genuine cross-cultural differences. In wine review adaptation, we even need to adapt the flavor descriptors. As is common in text generation tasks, we first adopt reference-based automatic evaluation metrics. Moreover, considering reference-based metrics are often unreliable for subjective tasks, we also conduct human evaluations.
\subsection{Automatic Evaluation}
We use four metrics to assess the similarity between the generated and reference reviews. We use two lexical-based metric: BLEU \cite{papineni-etal-2002-bleu}, a precision metric based on token n-gram which emphasizes precision and commonly used in machine translation evaluation and METEOR \cite{banerjee-lavie-2005-meteor}, which combines precision and recall while incorporating linguistic features such as stemming and synonymy to provide a more comprehensive evaluation; one contextual-embedding based metric: BERTScore \cite{bert-score}, based on cosine similarity of contextualized token embeddings and capture deep semantic matching; one hybrid-based metric: Beer \cite{stanojevic-simaan-2014-beer}, based on multi-feature fusion regression indicator, which combines syntactic and semantic features to automatically evaluate translation quality through regression model learning. 
\subsection{Human Evaluation}
While automatic metrics provide quantifiable results, they rely on fixed reference sets, which may lack cultural relevance. To address this, we introduce seven human evaluation criteria applied to the test set.

(1) \textbf{Grammar}:  The generated reviews are grammatically correct and fluent; (2) \textbf{Faithfulness of Information}:  The content accurately reflects the original input without introducing false or misleading information; (3) \textbf{Faithfulness of Style}: The output preserves the original tone, register, and formality without altering the intended voice; (4) \textbf{Overall quality}: The reviews are coherent, contextually appropriate, and align with the intended tone and style; (5) \textbf{Cultural Proximity}: The generated reviews use familiar terms and expressions that resonate with the target culture. For example, black currant was replaced by Chuanbei loquat paste, cough syrup, and hawthorn cake in the Chinese localised aroma wheel \cite{jin2022chinese};
(6) \textbf{Cultural Neutrality}: Maintains neutrality to avoid provoking negative perceptions or reactions from the target culture consumers. For example, 'earthy' can be a positive descriptor for wine in the West. However, when translated into Chinese, '\begin{CJK}{UTF8}{gbsn}土味\end{CJK}' often implies a dirty or unrefined flavor. A more elegant term like ‘\begin{CJK}{UTF8}{gbsn}泥土气息\end{CJK}’ (aroma of soil) is often preferred; (7) \textbf{Cultural Genuineness}: Preserves the quality of the original descriptor without altering its meaning, ensuring authenticity. For example, clove, violet, and saffron have no suitable local descriptors with similar olfactory characteristics.

Our evaluation was done by five people fluent in both Chinese and English, including four master students from various majors \footnote{The annotators are from Computer Science, Food Science, and Data Science.} and a Master of Wine. Before the evaluation process, we performed preliminary testing on 40 samples and used Pearson correlation to calculate their understanding of different metrics, which proved that these metrics are easy for people to evaluate.

\section{Cross-Cultural Reviews Analysis}
Here, we perform several analyses to answer \textbf{RQ1} and \textbf{RQ2}, highlighting the significant differences between Chinese and Western wine reviews and emphasizing the need for cultural adaptation.
\subsection{Lexical difference analysis}
To answer \textbf{RQ1}, we first manually extracted detailed flavor descriptors from 250 Chinese and 250 Western wine reviews and map detailed flavor descriptors into three hierarchical layers: \textbf{Aroma Families} (broad categories such as fruits, flowers, and spices), \textbf{Aroma Subfamilies} (more specific groups like citrus fruits, red berries, and dried herbs), and \textbf{Exact Aromas} (precise descriptors such as lemon, raspberry, or thyme)—leveraging the Wine Aroma Wheel from Aromaster\footnote{\url{https://aromaster.com/}}, with Chinese-localized descriptors proposed by \citep{jin2022chinese}. We found that over 300 unique descriptors were identified in the corpus, although they totally offer about 100 flavor descriptors.  To facilitate systematic cross-cultural comparison, we constructed a culturally aligned inventory of 92 standardized descriptors from the merged wheel, which includes flavors mentioned at least five times in the selected reviews. We attribute the variation in descriptor usage to individual author preferences rather than broadly shared sensory conventions. An example of how we applied this method can be found in Appendix \ref{sec:Jaccard}. Based on this mapped inventory, we then compute the inner- and outer-group Jaccard similarities\footnote{“Outer” refers to comparisons between Chinese and Western reviews, while “Inner” refers to comparisons within Western reviews.} over these Aroma Hierarchies.

\begin{table}[h]
\centering
\resizebox{\linewidth}{!}{ 
\begin{tabular}{c|c|c|c}
\toprule
          &  Exact Aromas    &  Aroma Subfamilies    &  Aroma Families    \\ \midrule
Inner Sim & 0.14 & 0.25 & 0.48 \\ \midrule
Outer Sim & 0.08 & 0.18 & 0.40\\ \bottomrule
\end{tabular}
}
\caption{Inner- and Outer-Group Jaccard Similarity Across Aroma Hierarchies}
\label{tab:Jaccard}
\end{table}

From Table \ref{tab:Jaccard}, we observe that at the Exact-Aroma level the average inner-group Jaccard similarity is 0.14, whereas the outer-group score falls to 0.08, revealing scant lexical overlap across cultures. At the Aroma-Subfamily level, the outer similarity improves to 0.18, yet the gap to the inner score (0.25) persists, confirming noticeable divergence. Even at the most abstract tier—Aroma Families—the outer similarity climbs only to 0.40, still trailing the inner benchmark (0.48) and demonstrating that a sizeable cultural difference remains despite hierarchical mapping. These results are statistically confirmed Appendix~\ref{sec:sig_Test}.

Overall, however, cross-cultural overlap remains modest even at the Aroma Families level, which we attribute to contrasting reviewing conventions: some critics mention only the dominant notes of complex wines, whereas others enumerate every perceivable aroma. And in both cultures, people still have their own tendencies toward specific flavor descriptors. The consistently lower outer-group scores highlight a clear need for cultural adaptation: without targeted lexical substitution, flavor descriptions that feel familiar to one readership may remain opaque—or misleading—to another. A detailed case study of lexical similarities and cultural differences for a representative wine (\textit{Chateau Valandraud 2016}) is provided in Appendix~\ref{sec:cul_diff}.

\subsection{Semantic difference analysis}
\begin{figure}[ht]
    \centering
    \includegraphics[width=0.9\linewidth]{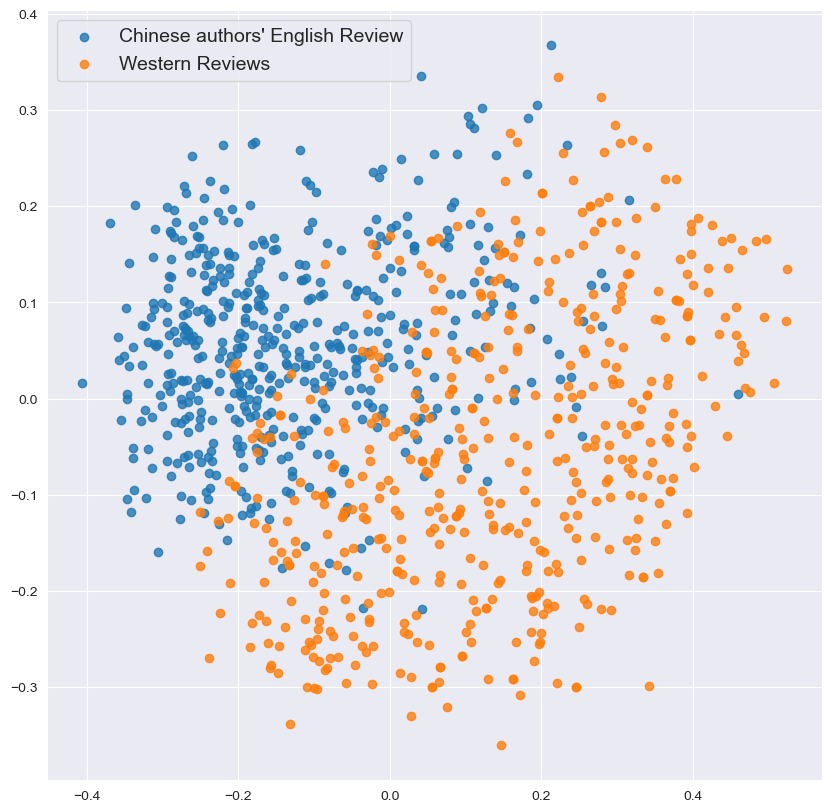}
    \caption{PCA-reduced embeddings: Blue dots show Chinese Authors' English reviews, and orange dots show Western reviews}
    \label{fig:embedding}
\end{figure}

To further analyze semantic differences, we perform PCA on the review embeddings, which are extracted from the encoder layer of LaBSE \citep{feng2022languageagnosticbertsentenceembedding}.
To mitigate distortion from dimensionality reduction and balance sample sizes, we uniformly sample 500 instances from each category. In addition, to minimize language-specific bias, we focus on English reviews authored by Chinese reviewers. As shown in Figure \ref{fig:embedding}, the two embedding clouds are almost separated in the reduced space, indicating substantial differences in the global structure and overall semantic distribution of the two review groups. When we instead contrast native Chinese-language reviews with English-language reviews, the separation widens even further, underscoring the strong confounding effect of surface language. The result can be seen in Appendix \ref{sec:Vis}. Synthesizing evidence from both the embedding-level and lexical level, Chinese and Western reviews remain essential differences even under strict controls, further reinforcing the need for cultural adaptation. 

\subsection{Cross-Cultural Comprehension Analysis}
To answer \textbf{RQ2}, we randomly select 40 reviews from each culture and let three annotators label the initial untranslated reviews on Cultural Proximity and Cultural Neutrality.\footnote{Inner = ratings from annotators of the same culture; Outer = ratings from annotators of the different culture.}

\begin{table}[h]
\centering
\resizebox{\linewidth}{!}{ 
\begin{tabular}{c|c|c|c|c}
\toprule
  Culture        &  Inner C-P    &  Outer C-P  &   Inner C-N   &  Outer C-N   \\ \midrule
Chinese & 4.9 & 4.5 & 5.3 & 4.7 \\ \midrule
Western & 6.2 & 5.6 & 6.7 & 5.3 \\ \bottomrule
\end{tabular}
}
\caption{Cross-Cultural Ratings of Cultural Proximity and Neutrality in Chinese and Western Reviews}
\label{tab:RQ2}
\end{table}

As shown in Table~\ref{tab:RQ2}, annotators tend to give higher scores to reviews from their own culture. Additionally, Western annotators consistently assign higher scores than Chinese annotators, suggesting that Western wine reviews may be more difficult for Chinese readers to understand, compared to how Western readers perceive Chinese wine reviews.

\section{Experiment}
To comprehensively assess the efficacy of LLM translations in understanding wine and flavors, we compare various prompting strategies on tuning-free LLMs, alongside evaluations of an open-source Machine Translation model.
\subsection{Experimental Setup}
\textbf{Prompting LLMs.} Based on the exceptional performance of multilingual LLMs in zero-shot translation. We explore their ability on translating wine reviews and flavor adaptation.

We compare the performance of five diverse, state-of-the-art LLMs on this task. We test Llama-3.1-8B-Instruct \cite{grattafiori2024llama}, Mistral-7B-Instruct-v0.3 \cite{jiang2023mistral7b},Phi-3.5-mini-instruct \cite{abdin2024phi3technicalreporthighly}, and ChatGPT-4o \cite{openai2024gpt4ocard} and two use more Chinese training data models: Qwen2.5-7B-Instruct \cite{qwen2.5}, GLM4-9b \cite{glm2024chatglm}. We used Cultural Prompt in Table \ref{tab:Prompt}.

\textbf{Multilingual machine translation model:} 
We use the state-of-the-art NLLB-200-3.3B model \cite{nllbteam2022languageleftbehindscaling} for accurate, context-aware multilingual translation in our experiments.

%% file: latex/Results.tex
\section{Results and Analysis}
Our analysis includes five parts: 1) automatic evaluation between different models; 2) fine-grained human evaluation on a subset of Chinese-English translation bidirectional; 3) Correlation of automatic metrics with humans; 4) Different prompting strategy evaluation comparison; 5) Quantitative analysis for some specific concepts

\subsection{Overall Automatic Evaluation}
\begin{table}[!ht]
    \centering
    \resizebox{\linewidth}{!}{ 
        \begin{tabular}{c c c c c c c}
            \hline
            \textbf{Models} & \textbf{BLEU} & \textbf{METEOR} & \textbf{B-Sc} & \textbf{BEER} & \textbf{\#Tok} & \\
            \hline
            \multicolumn{7}{c}{\textbf{Chinese $\rightarrow$ English}} \\
            \hline
            ChatGLM4 & \underline{\textbf{18.7}} & 45.7 & 86.7 & 51.5 & 65.8 & \\
            Phi   & 10.9 & 40.6 & 90.0 & 47.5 & 75.6 & \\
            Qwen2.5 & 15.6 & 46.3 & \underline{\textbf{91.2}} & 52.7 & 69.2 & \\
            Mistral & 5.2 & 36.4 & 87.9 & 34.1 & 61.8 & \\
            Llama3.1 & 11.4 & 35.2 & 88.0 & 44.3 & 54.0 & \\
            NLLB & 12.0 & 36.8 & 88.7 & 48.1 & 64.8 & \\
            ChatGPT & 18.4 & \underline{\textbf{48.5}} & \underline{\textbf{91.2}} & \underline{\textbf{53.4}} & 78.5 & \\
            \hline
            \multicolumn{7}{c}{\textbf{English $\rightarrow$ Chinese}} \\
            \hline
            ChatGLM4 & 8.9 & 31.9 & 86.6 & 26.5 & 130.5 & \\
            Phi   & 4.8 & 25.8 & 89.8 & 22.5 & 94.7 & \\
            Qwen2.5 & 12.3 & 36.4 & \underline{\textbf{90.6}} & 29.3 & 85.7 & \\
            Mistral & 2.0 & 20.1 & 89.3 & 17.0 & 56.1 & \\
            Llama3.1 & 9.9 & 31.9 & 87.3 & 28.1 & 89.5 & \\
            NLLB & 3.8 & 18.3 & 87.6 & 21.4 & 96.8 & \\
            ChatGPT & \underline{\textbf{16.4}} & \underline{\textbf{41.1}} & 90.2 & \underline{\textbf{34.0}} & 90.3 & \\
            \hline
        \end{tabular}
    }
    \caption{Automated evaluation results on the test sets using reference-based metrics: BLEU, METEOR, B-Sc(BERTScore) and BEER. Higher scores indicate better performance on all metrics.}
    \label{tab:auto_metrics}
\end{table}

\noindent
For Chinese-to-English translation, ChatGLM4 achieved the highest BLEU score, suggesting strong lexical matching, while ChatGPT excelled in BERTScore and BEER, indicating superior semantic alignment with human-written translations. For English-to-Chinese translation, ChatGPT led in BLEU and METEOR, and Qwen2.5 again outperformed in BERTScore, reinforcing its strength in preserving meaning. Notably, ChatGLM4 struggled with BLEU and METEOR in this direction but maintained competitive BERTScore values, suggesting it prioritizes semantic coherence over strict lexical overlap. Additionally, the NMT system NLLB still remains highly competitive. Overall, ChatGPT and Qwen2.5 consistently demonstrated high semantic quality across both directions, while other models exhibited strengths in specific metrics, reflecting differing optimization objectives. These results highlight the inherent trade-offs between fluency, lexical fidelity, and semantic preservation across models. More importantly, they underscore that reference translations are not the sole "correct" adaptations, as translation quality is inherently subjective. This reinforces the need for a nuanced evaluation framework that accounts for cultural context, linguistic variation, and domain-specific preferences to better capture real-world translation quality.


\subsection{Human Evaluation}
\begin{table}[ht]
    \centering
    \resizebox{\linewidth}{!}{ 
        \begin{tabular}{c c c c c c c c c}
            \hline
            \textbf{Models} & \textbf{F-I} & \textbf{F-S} & \textbf{Gr} & \textbf{O-Q} & \textbf{C-P} & \textbf{C-G} & \textbf{C-N}\\
            \hline
            \multicolumn{7}{c}{\textbf{Chinese $\rightarrow$ English(n=42)}} \\
            \hline
            ChatGLM4 & 5.6 & 4.8 & 5.3 & 5.0 & 6.6 & \underline{\textbf{6.4}} & 6.2 \\
            Phi   & 4.8 & 5.4 & 6.0 & 4.8 & 6.6 & 6.3 & \underline{\textbf{6.4}} \\
            Qwen2.5 & 5.8 & \underline{\textbf{6.0}} & 5.8 & 5.5 & 6.5 & 6.3 & \underline{\textbf{6.4}} \\
            Mistral & 4.8 & 5.7 & 5.8 & 4.8 & \underline{\textbf{6.7}} & \underline{\textbf{6.4}} & 6.3\\
            Llama3.1 & 5.3 & 5.6 & \underline{\textbf{6.2}} & 5.3 & 6.6 & 6.2 & 6.3 \\
            Human & 5.5 & 5.3 & 5.5 & 5.3 & 6.3 & 6.3 & 6.2\\
            ChatGPT & \underline{\textbf{5.9}} &  \underline{\textbf{6.0}} &  5.7 &  \underline{\textbf{5.8}} &  6.5 &  6.1 &  6.3 \\
            \hline
            \multicolumn{7}{c}{\textbf{English $\rightarrow$ Chinese(n=82)}} \\
            \hline
            ChatGLM4 & 5.5 & 5.5 & 4.2 & 4.7 & 5.6 & 5.5 & 5.3\\
            Phi   & 4.1 & 4.5 & 3.6 & 3.7 & 5.6 & 4.4 & 5.3\\
            Qwen2.5 & 5.7 & 5.9 & 5.5 & 5.6 & 6.0 & 5.5 & \underline{\textbf{5.8}}\\
            Mistral & 3.8 & 4.4 & 2.8 & 3.1 & 4.8 & 4.6 & 4.7\\
            Llama3.1 & 4.5 & 5.0 & 4.4 & 4.6 & 5.9 & 4.8 & 5.7\\
             Human & 5.1 & 5.8 & 5.2 & 5.0 & 5.8 & \underline{\textbf{5.7}} & \underline{\textbf{5.8}}\\
             ChatGPT & \underline{\textbf{6.2}} &  \underline{\textbf{6.1}} &  \underline{\textbf{5.8}} &  \underline{\textbf{5.9}} &  4.5 &  6.1 &  5.7 \\
            \hline
        \end{tabular}
    }
    \caption{Human evaluation results on the selected test sets: average for each method and metric, ranging from 1 to 7. F-I, F-S, Gr, O-Q, C-P, C-G and C-N representing Faithful of Information, Faithful of Style, Grammar, Overall Quality, Cultural Proximity, Cultural Genuineness and Cultural Neutrality respectively}
    \label{tab:human_eval}
\end{table}

Table \ref{tab:human_eval} presents the results of human evaluation across multiple dimensions. Notably, NLLB was excluded from human evaluation due to its lack of cultural adaptation capabilities.

For English-to-Chinese translation, ChatGPT leads across most metrics, even surpassing human translations in faithfulness, grammar, and overall quality. Human translations score highest in Cultural Genuineness, reflecting their ability to preserve nuanced expressions. Qwen2.5 also remains competitive, balancing fluency and cultural adaptation.

For Chinese-to-English translation, ChatGPT again ranks highest in faithfulness and overall quality, while Mistral outperforms even human translations in Cultural Proximity and Genuineness, suggesting a stronger emphasis on natural, idiomatic English.

These results highlight trade-offs between literal accuracy and cultural adaptation. While human translations excel in cultural authenticity, LLMs like Qwen2.5 demonstrate strong faithfulness, and Mistral prioritizes fluency in the target language. This underscores the importance of context-aware evaluation that considers both linguistic accuracy and cultural nuances.

For the feedback from the Master of Wine, we still find that Model ChatGLM has correct vocabulary but has poor grammar, also very literal. Phi has common wrong translations, but has fixed wine tasting vocabulary. Llama is the most metaphorical and tend to reword stuff. Mistral just skips proper nouns but eloquent.
ChatGLM, ChatGPT and Human translation are best. Human translation and ChatGPT are maybe too literal (correct but boring), ChatGLM are super eloquent.

\textbf{GPT-4o and Human Evaluation Diverge.} Table \ref{tab:LLM_Judge} shows the results evaluated by both GPT-4o and human annotators. Specifically, we use all the human evaluation test cases for GPT evaluation. The results show that GPT and Human evaluation scores are not strongly correlated, and GPT has a higher tolerance than humans, especially for the English $\rightarrow$ Chinese direction. This discrepancy suggests that GPT-4o may have a different interpretation of translation quality compared to human evaluators. This discrepancy is particularly evident in culturally relevant metrics. The prompts we used are shown in Appendix~\ref{sec:Prompt}.
\begin{table}[ht]
    \centering
    \resizebox{\linewidth}{!}{ 
        \begin{tabular}{c c c c c c c c c}
            \hline
            \textbf{Methods} & \textbf{F-I} & \textbf{F-S} & \textbf{Gr} & \textbf{O-Q} & \textbf{C-P} & \textbf{C-G} & \textbf{C-N}\\
            \hline
            \multicolumn{7}{c}{\textbf{Chinese $\rightarrow$ English}} \\
            \hline
            Human-Eval & 5.36 &5.58 &5.91 &5.19 &6.86 &6.31 &6.36  \\
            GPT4o-Eval & 5.79 &5.24 &6.47 &5.77 &5.23 &5.4 &6.44 \\
            \hline
            \multicolumn{7}{c}{\textbf{English $\rightarrow$ Chinese}} \\
            \hline
            Human-Eval & 4.57 &5.02 &4.44 &4.13 &5.25 &5.12 &5.0 \\
            GPT4o-Eval & 5.76 &5.36 &6.2 &5.71 &5.39 &5.58 &6.43 \\
            \hline
        \end{tabular}
    }
    \caption{GPT-4o v.s. Human in Human evaluation metrics (Average over All Human Test Cases)}
    \label{tab:LLM_Judge}
\end{table}

\subsection{Correlation of automatic metrics with humans}
\begin{table}[ht]
    \centering
    \resizebox{\linewidth}{!}{ 
        \begin{tabular}{c c c c c c}
            \hline
             & \textbf{BLEU} & \textbf{METEOR} & \textbf{B-Sc} & \textbf{BEER} &\\
            \hline
            \multicolumn{6}{c}{\textbf{Chinese $\rightarrow$ English}} \\
            \hline
            \textbf{F-I} & 0.2536* & 0.1892 & 0.3000* & 0.2442*\\
            \textbf{F-S}  & 0.0053 & 0.0075 & 0.0204 & 0.0113\\
            \textbf{Gr} & -0.0296 & -0.0096 & 0.0033 & -0.0478\\
            \textbf{O-Q} & 0.2191* & 0.1847* & 0.2061* & 0.2015*\\
            \textbf{C-P} & -0.070 & -0.0177 & -0.0540 & -0.0961\\
            \textbf{C-G} & 0.1131 & 0.0625 & 0.0621 & 0.1131\\
            \textbf{C-N}  & -0.1166 & -0.0841 & -0.0166 & -0.0876\\
            \hline
            \multicolumn{6}{c}{\textbf{English $\rightarrow$ Chinese}} \\
            \hline
            \textbf{F-I} & 0.4079* & 0.2788* & 0.4080* & 0.2946*\\
            \textbf{F-S}  & 0.3723* & 0.3560 & 0.3769* & 0.3904*\\
            \textbf{Gr} & 0.2819* & 0.2788* & 0.3530* & 0.2946*\\
            \textbf{O-Q} & 0.3526* & 0.3123* & 0.3742* & 0.3557*\\
            \textbf{C-P} & 0.1408 & 0.1524 & 0.2609* & 0.1553\\
            \textbf{C-G} & 0.3134* & 0.3170* & 0.3942* & 0.3170*\\
            \textbf{C-N}  & 0.1334 & 0.1357 & 0.2389* & 0.1516\\
            \hline
        \end{tabular}
    }
    \caption{Kendall correlation of human evaluation results with automatic metrics. Statistically significant correlations are marked with *, with a confidence level of $\alpha = 0.05$ before adjusting for multiple comparisons using the Bonferroni correction}
    \label{tab:correlation}
\end{table}
To evaluate the reliability of automatic metrics for wine review adaptations, we analyze their correlation with human evaluations across seven metrics using Kendall correlation, the WMT22 meta-evaluation standard \cite{freitag-etal-2022-results}.

As illustrated in Table \ref{tab:correlation}, the correlation between human evaluation results and automatic metrics varies across different translation directions and evaluation criteria. For Chinese → English, F-I (Faithful of Information) and O-Q (Overall Quality) exhibit the strongest correlations, particularly with BLEU, B-Sc, and BEER, suggesting that these metrics align well with human judgments in assessing fluency and overall translation quality. On the other hand, for English-Chinese, the correlations are generally stronger across all metrics. Notably, F-I, F-S (Faithful of Style), and O-Q display significant correlations with multiple metrics, particularly B-Sc and BEER, indicating that these metrics are relatively more reliable for evaluating fluency and intelligibility in English-to-Chinese translations. 

C-G (Cultural Genuineness) also achieves strong correlations, indicating that automatic metrics can reliably assess translation accuracy. However, this does not necessarily reflect the cultural adaptability of the translation. However, C-N(Cultural Neutrality) and C-P(Cultural Proximity) remain weakly correlated, revealing that automatic metrics still fall short in capturing deeper cultural nuances, emphasizing the need for human evaluation. Notably, correlations for English→Chinese generally exhibit greater strength than Chinese→English. This discrepancy is likely due to most wine reviews being written from a predominantly Western reviewer's perspective.

\subsection{Prompting Strategy Evaluation}
\begin{table}[]
    \centering
    \begin{subtable}[t]{\linewidth}
        \centering
        \resizebox{\linewidth}{!}{ 
            \begin{tabular}{c c c c c c c c}
                \hline
                \multicolumn{7}{c}{\textbf{Human Evaluation}} \\
                \hline
                \textbf{Methods} & \textbf{F-I} & \textbf{F-S} & \textbf{Gr} & \textbf{O-Q} & \textbf{C-P} & \textbf{C-G} & \textbf{C-N}\\
                \hline
                Direct Translation & 6.38 &  5.94 &  5.56 &  5.69 &  4.38 &  6.31 &  5.81 \\
                Cultural Prompt & 6.24 &  6.12 &  5.83 &  5.94 &  4.5 &  6.12 &  5.74 \\
                Detailed Cultural Prompt & 5.75 &  5.94 &  5.75 &  5.56 &  4.88 &  5.69 &  5.69 \\
                Self-Explanation & 4.94 &  5.75 &  5.88 &  5.5 &  5.75 &  4.56 &  6.62 \\
                \hline
            \end{tabular}
        }
    \end{subtable}

    \begin{subtable}[t]{\linewidth}
        \centering
        \resizebox{\linewidth}{!}{ 
            \begin{tabular}{c c c c c }
                \hline
                \multicolumn{5}{c}{\textbf{Automatic Evaluation}} \\
                \hline
                \textbf{Methods} & \textbf{BLEU} & \textbf{METEOR} & \textbf{B-Sc} & \textbf{BEER} \\
                \hline
                Direct Translation & 16.5 & 41.6 & 91.4 & 28.3 \\
                Cultural Prompt & 15.9 & 41.0 & 91.3 & 28.3 \\
                Detailed Cultural Prompt & 14.4 & 38.6 & 91.0 & 27.6 \\
                Self-Explanation & 13.7 & 37.0 & 90.9 & 27.6 \\
                \hline
            \end{tabular}
        }
    \end{subtable}
    \caption{ Evaluation of different strategies by GPT4o on English-Chinese translations.}
    \label{tab:Prompt_Compare}
\end{table}
With LLM-based machine translation advancing, integrating free-form external knowledge offers new opportunities to enhance translation quality. We compare different prompting strategies on ChatGPT for English-to-Chinese translation, including Direct Translation, Cultural Prompt, Detailed Cultural Prompt, and Self-Explanation. Table \ref{tab:Prompt} lists the specific prompts.

As is shown in Table \ref{tab:Prompt_Compare}, Direct Translation is the best-performing method in Faithful of Information and automatic metrics. Cultural Prompting improves cultural accuracy slightly but does not significantly enhance overall translation quality. We attribute this to the high proportion of culture-agnostic sensory terms. Self-Explanation has the worst faithful scores but is rated the best in Cultural Neutrality, making it a trade-off strategy for culturally rich contexts. This trade-off suggests that translation strategies need to balance faith, accuracy, and cultural adaptability, depending on the intended use case.

%% file: latex/Conclusion.tex
\section{Conclusion}
In this work, we studied cross-cultural adaptation of wine reviews, introducing CulturalWR, a dataset of paired Chinese and English reviews, and evaluating LLM-based adaptation methods. Our results show that LLMs can consider cultural nuances but face challenges in maintaining detail and consistency in flavor descriptions. We also assessed adapted flavor similarities to gauge LLMs’ understanding of wine descriptors.
Beyond wine reviews, our findings have broader implications for cross-cultural communication in the wine industry, aiding wineries and retailers in tailoring descriptions to international audiences. This could enhance global wine marketing while preserving cultural authenticity. Moreover, our work highlights AI's potential in gastronomy, paving the way for research into AI-assisted flavor profiling and food pairing.
Future work includes refining adaptation models for coherence, integrating multimodal data, and using user feedback to enhance AI-generated adaptations, bridging cultural gaps in wine appreciation.

%% file: latex/Limitation.tex
\section{Limitation}
Cultural adaptation in wine reviews has great potential to aid consumer decisions and promote wines globally. However, several challenges remain. A key limitation is the dataset size and diversity. Our study includes fewer than 5,000 Chinese reviews, primarily from three professional critics, raising concerns about representativeness. 

Additionally, while our dataset contains reviews in multiple languages (German, Portuguese, French, Dutch, Italian, and Spanish), we focused solely on Chinese and English due to resource constraints. Expanding multilingual analysis could offer further insights. Another constraint lies in evaluation prompts. Our analysis relies on four prompts, which may not fully capture how models handle cultural adaptation. Broader prompt variations and real-world user inputs could improve evaluation robustness. 

Moreover, our quality assessment is based on the Wine Aroma Wheel and prior sensory adaptation research \cite{jin2022chinese}, but it lacks independent verification of adaptation accuracy. Future work could incorporate human or expert evaluations for a more comprehensive assessment. Finally, wine reviews are inherently subjective, influenced by personal preferences and sensory perceptions. While we strive to minimize bias, eliminating subjectivity entirely remains challenging. Addressing this may require larger, more diverse datasets and structured sensory evaluation frameworks in future research.

Finally, while the broader literature acknowledges that current LLMs lack reliable mechanisms for on-the-fly cultural adaptation, our work highlights concrete failure cases in the wine domain. Solving this challenge will likely require automatic detection of culturally opaque descriptors and culture-aware fine-tuning. We leave the design of such corrective mechanisms to future work.

%% file: latex/Appendix.tex
\section{French and Spanish Character Conversion Table}
\label{sec:Conversion}
This section shows the Character Conversion Table we have used to convert some characters, shown in Table \ref{tab:conversion}.
\begin{table}[ht]
\centering
\resizebox{\linewidth}{!}{
\begin{tabular}{|c|c|c|c|}
\hline
\textbf{Original} & \textbf{Converted} & \textbf{Original} & \textbf{Converted} \\ \hline
à & a & â & a \\ \hline
ä & a & é & e \\ \hline
è & e & ê & e \\ \hline
ë & e & î & i \\ \hline
ï & i & ô & o \\ \hline
ö & o & ù & u \\ \hline
û & u & ü & u \\ \hline
ç & c & ÿ & y \\ \hline
æ & ae & œ & oe \\ \hline
À & A & Â & A \\ \hline
Ä & A & É & E \\ \hline
È & E & Ê & E \\ \hline
Ë & E & Î & I \\ \hline
Ï & I & Ô & O \\ \hline
Ö & O & Ù & U \\ \hline
Û & U & Ü & U \\ \hline
Ç & C & Ÿ & Y \\ \hline
á & a & í & i \\ \hline
ó & o & ú & u \\ \hline
ñ & n & Á & A \\ \hline
Í & I & Ó & O \\ \hline
Ú & U & Ñ & N \\ \hline
\end{tabular}}
\caption{French and Spanish Character Mapping to English}
\label{tab:conversion}
\end{table}

\section{Whole dataset of Other Languages}
It's the all the data we have collected, besides Chinese and English, we also collected some reviews written in German, Portuguese, French, Dutch, Italian and Spanish.
\label{sec:Dataset}
\begin{table}[H]
\centering
\resizebox{\linewidth}{!}{ 
\begin{tabular}{c|c|c}
\hline
                                & Numbers & Mean Tokens \\ \hline
CA Chinese Reviews & 4776    & 67.57       \\ 
CA English Reviews & 3227    & 74.25       \\ \hline
WA English Reviews & 16746  & 58.16       \\ \hline
WA German Reviews & 3341   & 50.73       \\ \hline
WA Portuguese Reviews & 161   & 82.15       \\ \hline
WA French Reviews & 480   & 135.63       \\ \hline
WA Italian Reviews & 40   & 44.35       \\ \hline
WA Spanish Reviews & 10   & 114.8       \\ \hline
\end{tabular}
}
\caption{Statistics of reviews. We count tokens with jieba text segmentation for Chinese and whitespace tokenization for other languages.}
\label{tab:whole_stats}
\end{table}

\section{Significance test}
\label{sec:sig_Test}
\begin{table}[h]
\centering
\resizebox{\linewidth}{!}{
\begin{tabular}{lccc}
\toprule
\textbf{Aroma Level} & \textbf{Mean Difference $\Delta$} & \textbf{95\% CI$_\text{bootstrap}$} & \textbf{Significance} \\
\midrule
Exact Aromas        & $-0.045$ & [$-0.050$, $-0.041$] & Significant \\
Aroma Subfamilies   & $-0.061$ & [$-0.068$, $-0.054$] & Significant \\
Aroma Families      & $-0.079$ & [$-0.090$, $-0.068$] & Significant \\
\bottomrule
\end{tabular}
}
\caption{Bootstrap Confidence Intervals and Mean Differences of Jaccard Similarities Across Aroma Hierarchies}
\label{tab:appendix-significance}
\end{table}

$\Delta$ represents the mean difference between outer-group and inner-group Jaccard similarity (Outer - Inner).  
95\% confidence intervals are computed via 10,000 bootstrap resamples.  
All intervals lie entirely below 0, indicating statistically significant differences in lexical similarity across cultural groups at all semantic levels.

\section{Embedding Visualization}
\label{sec:Vis}
\begin{figure}[ht]
    \centering
    \includegraphics[width=0.9\linewidth]{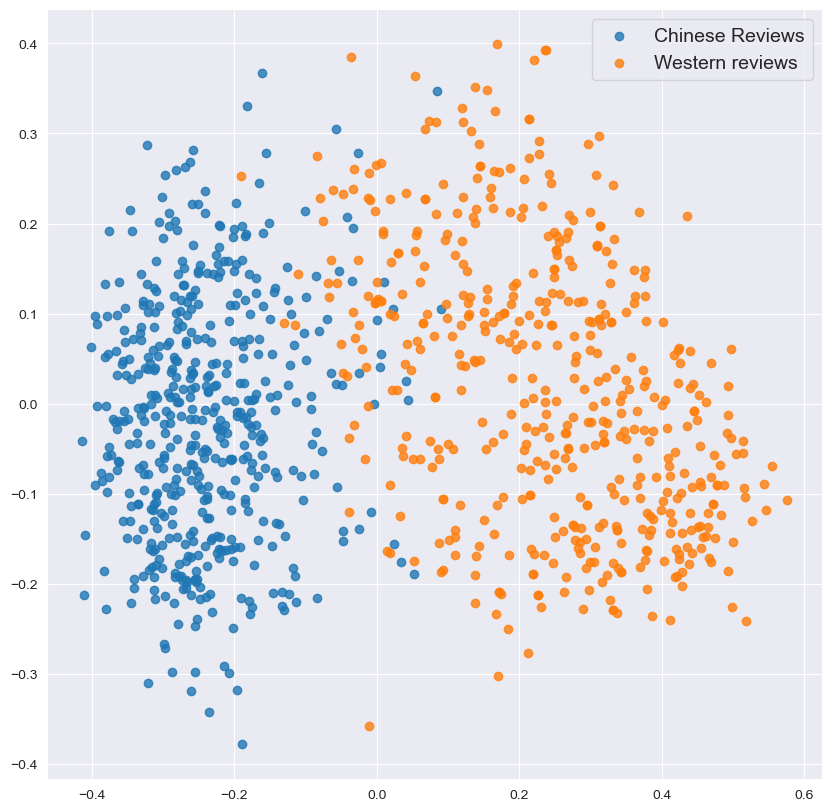}
    \caption{PCA-reduced embeddings: Blue dots show Chinese reviews, and orange dots show Western reviews}
    \label{fig:embedding_cn}
\end{figure}
The figure shows that PCA-reduced embedding of Chinese and Western reviews. Which widen the distance from \ref{fig:embedding}.

\section{Case Study: Cross-Cultural Differences in ‘Chateau Valandraud 2016’ Reviews}
\label{sec:cul_diff}
For the randomly selected wine \textit{Chateau Valandraud 2016}, our dataset contains two Chinese and four English reviews. 
Although the reviews vary in length and level of detail, we follow the lexical difference analysis described in Section~5.1. 
Specifically, we extract \textit{Aroma Families}, \textit{Aroma Subfamilies}, and \textit{Exact Aromas} from all reviews and compute the Jaccard similarities.

\begin{table}[h]
\centering
\resizebox{\linewidth}{!}{
\begin{tabular}{lccc}
\hline
 & Exact Aromas & Aroma Subfamilies & Aroma Families \\
\hline
Inner Sim & 0.05 & 0.16 & 0.53 \\
Outer Sim & 0.08 & 0.16 & 0.47 \\
\hline
\end{tabular}}
\caption{Jaccard Similarities between Chinese and English reviews of \textit{Chateau Valandraud 2016}.}
\label{tab:case-study-valandraud}
\end{table}

At the Exact Aroma level, the cross-cultural Jaccard similarity remains low, confirming that fine-grained vocabulary is highly discrete. 
We attribute the lower inner-group similarity compared to outer-group to contrasting reviewing conventions: 
some critics mention only the dominant notes of complex wines, whereas others enumerate every perceivable aroma.

As we move up to the Aroma Family level, the similarity between Chinese and English rises to 0.53 (inner group) and 0.47 (outer group). 
Although they are converging, the cross-cultural gap persists, consistent with the gap direction (0.48 vs. 0.40) observed in the macro corpus. 
This supports our paper's conclusion: in both cultures, reviewers maintain distinct tendencies toward specific flavor descriptors. 
The consistently lower outer-group scores further highlight the need for cultural adaptation.

Beyond the statistics, reviewers in both cultures select descriptors rooted in their cultural and sensory conventions: 
Chinese critics use denser concrete phrases and unique local metaphors such as ``yin--yang balance of tai chi,'' 
while English reviewers favor references common in Western tasting language (e.g., ``nutmeg,'' ``cinnamon,'' ``wood polish''). 
These patterns demonstrate that each cultural group gravitates toward familiar flavor expressions, reinforcing the need for adaptive translation.

\subsection*{Chinese Authors' Reviews}
\small
\begin{quote}
The wine integrates 90\% Merlot, 7\% Cabernet Franc and 3\% Cabernet Sauvignon. The appearance was dense and dark fuchsia. 
Exuding subtly introverted but precisely enticing, freshly aristocratic and fairly intact fragrances of stone, blackberry, loquat, underbush, vanilla and black chocolate. 
There was an exceedingly polished sense to the well-balanced flavours that oozed a large amount of austere and pure minerality together with succulent and vivid acid to impart the taste an uplift and supreme freshness that culminated in a compact and prolonged finale that was filled with talc-like tannins that reflected the sophistication of a winemaking master.

In a sweep of majestic elegance, ripe berry fruit, stony minerality, spicy oak and supple tannins strike a seamless harmony---much like the yin--yang balance of tai chi. 
The finish lands with an almost revelatory jolt, making it a wine to contemplate, sip after sip.
\end{quote}

\subsection*{Western Authors' Reviews}
\small
\begin{quote}
Incredible perfumes and beauty with ripe plum and berry character, as well as an array of spices, such as nutmeg and cinnamon. 
Full-bodied, firm and silky with beautifully polished tannins. Rendered and defined. A wine with lovely, classy nature and personality.

Expressive full, rich nose. Granular texture, grippy but with soft edges so you get a plump, almost plush feeling. 
Sophisticated and refined with dark, sultry black fruits.

Deep dark ruby, dense core, purple glints, faintly lighter at the rim. 
Intense new wood tones, vanilla, wood polish, smoke, cherry, coconut, and hazelnut. 
Immediately smooth onset, then very substantive, fleshy with astringency, barely contained green tones, both in acidity and tannins, surprisingly nervy. 
Pithy, seemingly driven structure.

Ripe, pure and layered nose, floral with violet, dark blackberry, raspberry, fine spiciness, very chalky and stony. 
Dense and concentrated palate, very intense and lingering with bright freshness, crushed berries, high concentration of tannin but very ripe and digest with an extremely long finish. 
Very pure, fresh and intense with perfect proportions.
\end{quote}
\normalsize

Collectively, these wine-specific results corroborate the robustness of our main conclusions while providing finer-grained insight into how cultural background shapes both what aromas are selected and how they are expressed.

\section{Prompt Used for Evaluation}
\label{sec:Prompt}
Table \ref{tab:Prompt} shows different prompt strategies we used for Prompting Strategy Evaluation.
\begin{table*}[ht]
\centering
\resizebox{\linewidth}{!}{
\begin{tabular}{cl}
\hline
Strategy                 & \multicolumn{1}{c}{Prompt}                                                 \\ \hline
Direct Translation       & Translate the following English wine reviews to Chinese: [Wine review]                                                  \\ \hline
Cultural Prompt          & Translate the wine reviews in Chinese, adapted to an Chinese-speaking consumer: [Wine review]                     \\ \hline
Detailed Cultural Prompt & \makecell[l]{Translate the provided English wine review into Chinese, so that it fits within Chinese wine culture and  to avoid using \\ any terms that might have negative connotations for Chinese consumers: [Wine review]}  \\  \hline
Self-Explanation         & \begin{tabular}[c]{@{}l@{}}User: Find flavor and aroma descriptions that are unfamiliar and uncomfortable for Chinese consumers: [Wine review]           \\ LLM: {[}Sentences{]}\\ User: Translate the wine review in Chinese, and for the unfamiliar and uncomfortable flavor and aroma,\\replace it with a more familiar and comfortable description for Chinese consumers.
\end{tabular} \\
\hline
\end{tabular}
}
\caption{Prompting strategy examples used for English $\rightarrow$ Chinese translation}
\label{tab:Prompt}
\end{table*}

\begin{table*}[ht]
\centering
\resizebox{\linewidth}{!}{
\begin{tabular}{cl}
\hline
Strategy                 & \multicolumn{1}{c}{Prompt}                                                 \\ \hline
Direct Translation       & Translate the following Chinese wine reviews to English: [Wine review]                                                  \\ \hline
Cultural Prompt          & Translate the wine reviews in English, adapted to an Western-speaking consumer: [Wine review]                     \\ \hline
Detailed Cultural Prompt & \makecell[l]{Translate the provided Chinese wine review into English, so that it fits within Western wine culture and to avoid using \\ any terms that might have negative connotations for Western consumers: [Wine review]}  \\  \hline
Self-Explanation         & \begin{tabular}[c]{@{}l@{}}User: Find flavor and aroma descriptions that are unfamiliar and uncomfortable for Western consumers: [Wine review]           \\ LLM: {[}Sentences{]}\\ User: Translate the wine review in English, and for the unfamiliar and uncomfortable flavor and aroma,\\replace it with a more familiar and comfortable description for Western consumers.
\end{tabular} \\
\hline
\end{tabular}
}
\caption{Prompting strategy examples used for Chinese $\rightarrow$ English translation}
\label{tab:Prompt_cn}
\end{table*}

Here shows the prompt we used for GPT evaluation:
\begin{verbatim}
    As a Western consumer, evaluate the 
    quality of wine review translation 
    from these dimensions, use seven-tier 
    scoring ranging from 1-7 : 7 means 
    Excellent, 6 means Very Good, 5 
    means Good, 4 means Fair, 3 means 
    Poor, 2 means Very Poor, 1 means 
    Nonsense: Grammar,Faithfulness of 
    Information,Faithfulness of Style,
    Overall quality,Cultural proximity
    - The generated reviews use 
    familiar terms and expressions that
    resonate with the target culture, 
    Cultural neutrality - Maintains 
    neutrality to avoid provoking 
    negative perceptions or reactions 
    from the target culture consumers, 
    Cultural Genuineness - Preserves 
    the quality of the original 
    descriptor without altering its 
    meaning, ensuring authenticity.
    Original: {original}
    Translation: {translation}
\end{verbatim}

\section{Grading scale rule for human evaluation}
\label{sec:human_eval}
Here shows the rule for the Seven-tier rating system:
\begin{enumerate}
    \item Excellent (rating: 7 points)\\ The translation is also highly matched in context and culture.
    \item Very good (rating: 6 points) \\ There may be slight (1-2) inaccuracy of vocabulary or inadequate reflection of some details, but it does not affect the overall understanding.
    \item Good (rating: 5 points) \\ There are individual mistranslations, but they will not seriously change the overall meaning.
    \item Fair (score: 4 points) \\ Mistranslations are obvious, which may cause some semantics to deviate from the original text.
    \item Poor (score: 3 points) \\ The overall translation quality is low and may mislead readers.
    \item Very poor (score: 2 points) \\ It is basically impossible to rely on the translation to understand the original text.
    \item Nonsense  (score: 1 point) \\ The translation is completely incoherent semantically and unable to convey the information which the original review tries to convey.
\end{enumerate}

\section{Human evaluation platform}
\label{sec:eval}
\begin{figure}[ht]
    \centering
    \includegraphics[width=0.48\textwidth]{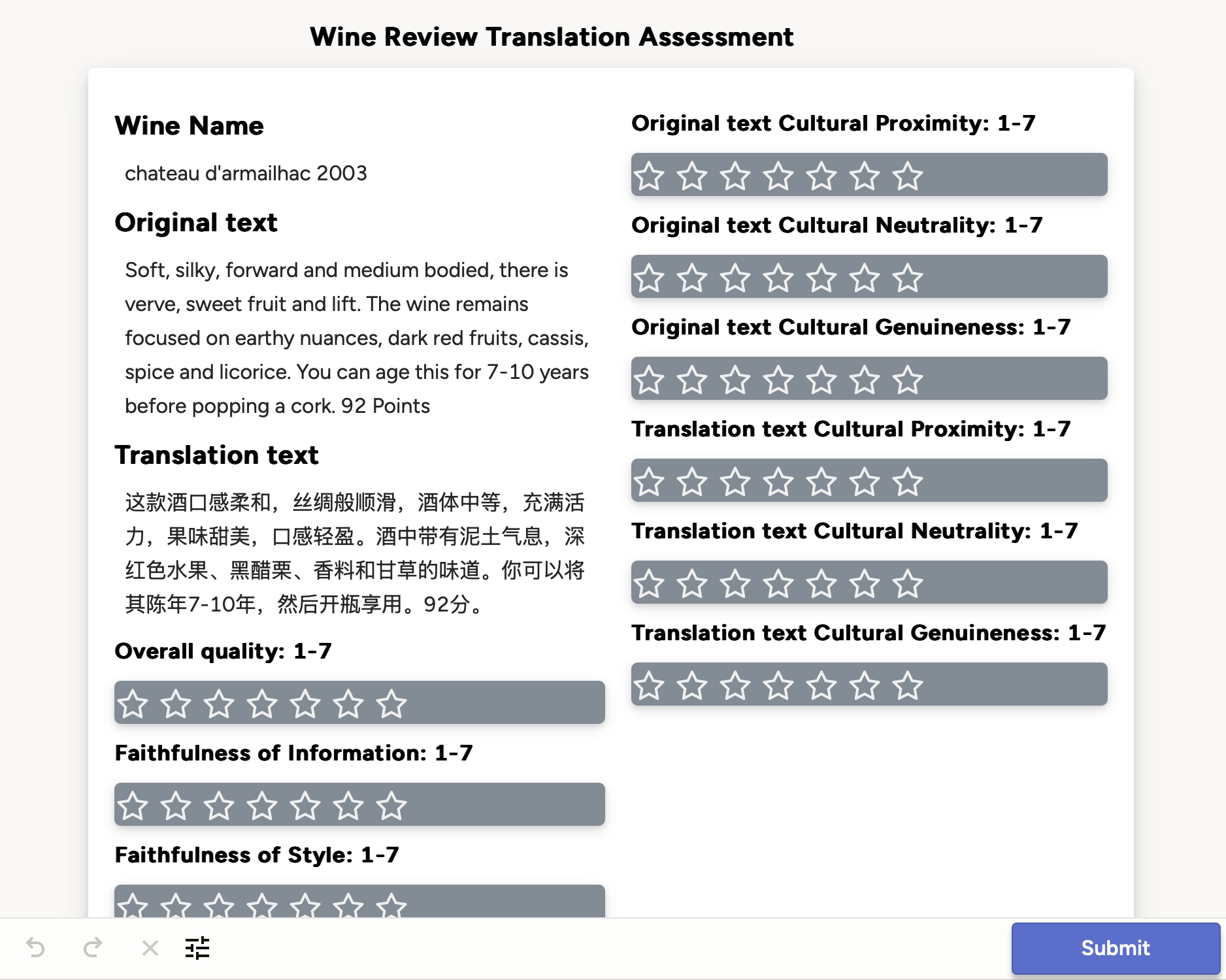}
    \caption{Screenshot from our human evaluation platform}
    \label{fig:screenshot}
\end{figure}

Figure \ref{fig:screenshot} shows a screenshot from our human evaluation platform, demonstrating the English to Chinese direction. Human need to evaluation the quality of Chinese translation

\section{Attributes of Whole dataset}
\label{sec:attributes}
\begin{figure*}[ht]
    \centering
    \begin{subfigure}[b]{0.48\linewidth}
        \centering
        \includegraphics[width=\linewidth]{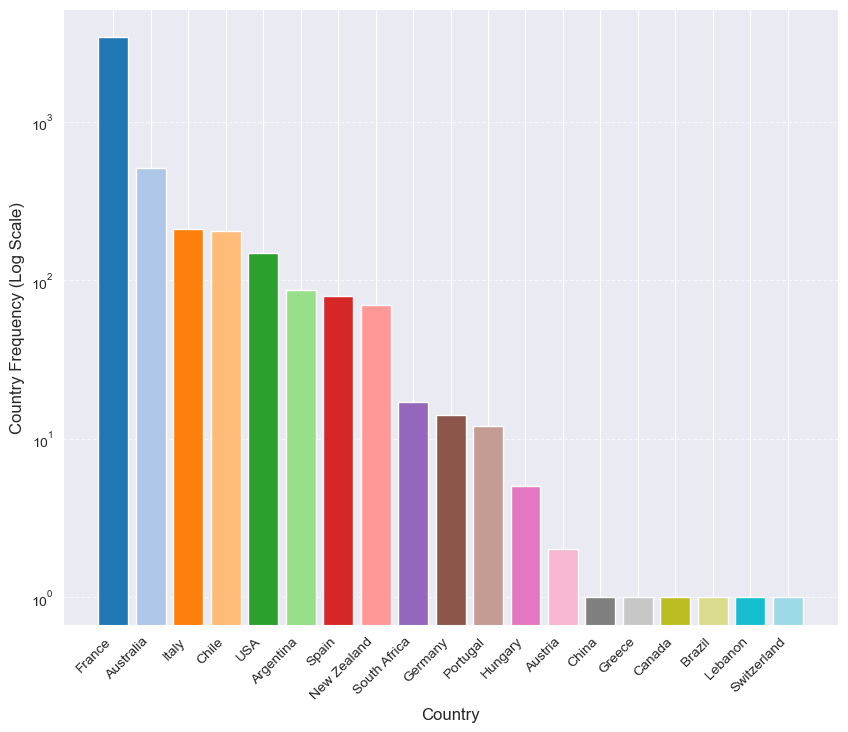}
        \caption{Country}
    \end{subfigure}
    \hfill
    \begin{subfigure}[b]{0.48\linewidth}
        \centering
        \includegraphics[width=\linewidth]{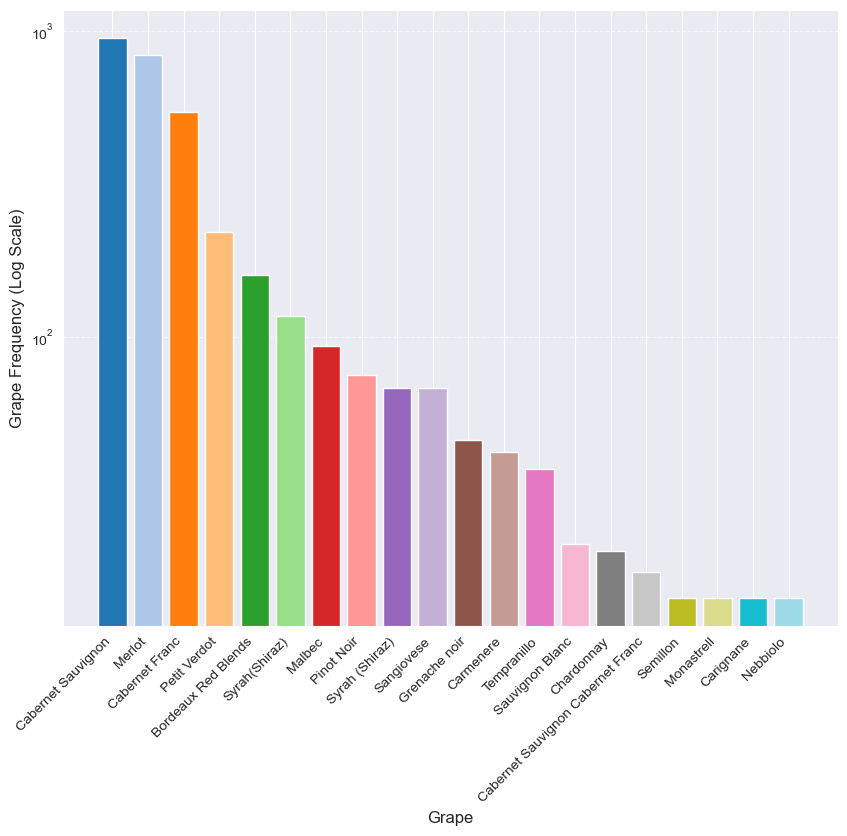}
        \caption{Grape}
    \end{subfigure}

    \begin{subfigure}[b]{0.48\linewidth}
        \centering
        \includegraphics[width=\linewidth]{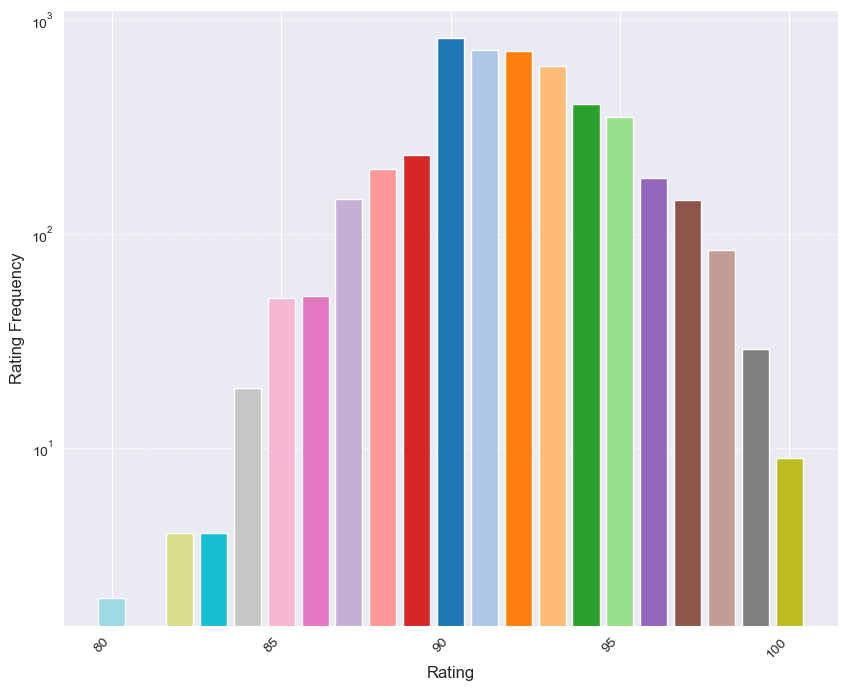}
        \caption{Rating}
    \end{subfigure}
    \hfill
    \begin{subfigure}[b]{0.48\linewidth}
        \centering
        \includegraphics[width=\linewidth]{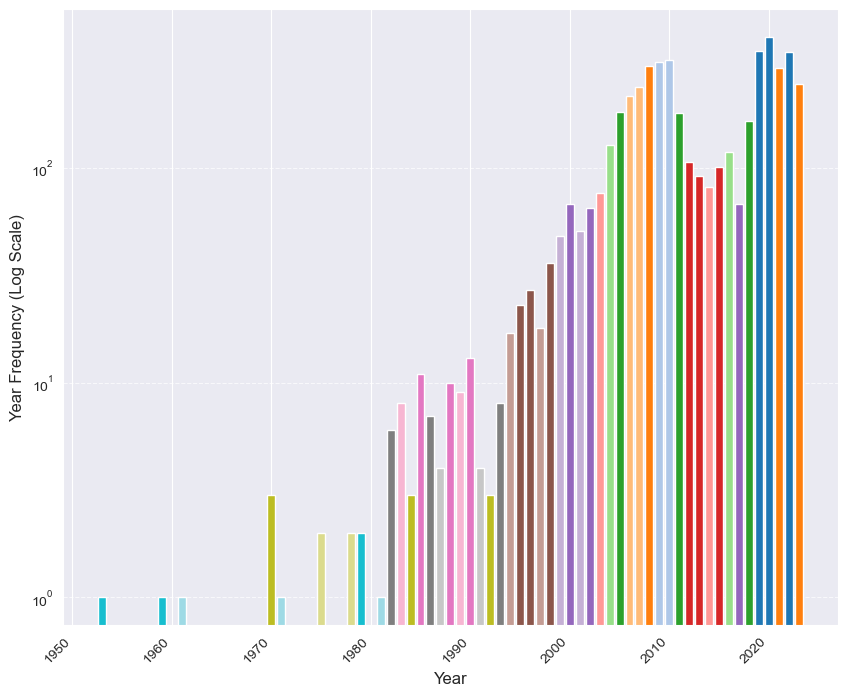}
        \caption{Year}
    \end{subfigure}

    \caption{Wine attributes.}
    \label{fig:attributes}
\end{figure*}

Figure \ref{fig:attributes} shows attributes counted in CulturalWR dataset.

\section{Experiment Settings}
\label{sec:setting}
The experiment settings of different models included in our paper are as follows:
\begin{enumerate}
    \item \textbf{NLLB} We use NLLB-200-3.3B\footnote{\url{https://huggingface.co/facebook/nllb-200-3.3B}} for our experiments. The beam is set as 4, and the length penalty is set as 1.0.
    \item \textbf{Pretrained LLMs} We used LLMs including Llama-3.1-8B-Instruct \footnote{\url{https://huggingface.co/meta-llama/Llama-3.1-8B-Instruct}}, Mistral-7B-Instruct-v0.3 \footnote{\url{https://huggingface.co/mistralai/Mistral-7B-Instruct-v0.3}}, Phi-3.5-mini-instruct\footnote{\url{https://huggingface.co/microsoft/Phi-3.5-mini-instruct}}, Qwen2.5-7B-Instruct\footnote{\url{https://huggingface.co/Qwen/Qwen2.5-7B-Instruct}}, GLM4-9b\footnote{\url{https://huggingface.co/THUDM/glm-4-9b}}. The sampling is set as True, leading to a multinomial sampling searching method. All settings are the same across different models.
    \item \textbf{ChatGPT} We used the latest version, GPT-4o-2024-11-20, through the ChatCompletion API provided by OpenAPI \footnote{\url{https://platform.openai.com/docs/guides/text-generation}}. For the generation, we set the parameters as default, for which the temperature is 1, top\_p is 1, and frequency\_penalty as 0.
\end{enumerate}

\section{Examples}
\label{sec:examples}
\begin{table*}[ht]
    \centering
    \renewcommand{\arraystretch}{1.3} 
    \begin{tabularx}{\textwidth}{|l|X|}
        \hline
        \textbf{Category} & \textbf{Content} \\
        \hline
        \textbf{Input (English)} & Aromas of iris, raspberry, camphor and Mediterranean scrub mingle with oak-driven spice on this 100\% Merlot. \\
        \hline
        \textbf{ChatGLM4} & \begin{CJK}{UTF8}{gbsn} 这款酒散发着\textcolor{blue}{紫罗兰}、\textcolor{blue}{树莓}、\textcolor{teal}{樟脑}和\textcolor{teal}{地中海灌木}的香气，100\%的梅洛。 \end{CJK} \\
        \hline
        \textbf{Phi} & \begin{CJK}{UTF8}{gbsn} 这100\%的\textcolor{red}{梅鹿酒}充满了\textcolor{blue}{茉莉花}、\textcolor{teal}{覆盆子}、\textcolor{brown}{茴香}和\textcolor{red}{地中海疏林}的香气，与橡木风味相结合的精髓。\end{CJK} \\
        \hline
        \textbf{Qwen2.5} & \begin{CJK}{UTF8}{gbsn}
            这款100\%梅洛红酒散发出\textcolor{teal}{鸢尾花}、\textcolor{teal}{覆盆子}、\textcolor{red}{薄荷}和\textcolor{teal}{地中海灌木丛}的香气，与橡木带来的香料味交织在一起。
        \end{CJK} \\
        \hline
        \textbf{Mistral} & \begin{CJK}{UTF8}{gbsn}
            这是一款\textcolor{red}{百分之一}的梅洛酒，浓郁而舒服，携带了\textcolor{brown}{芙蓉花}、\textcolor{blue}{莓果}、\textcolor{teal}{樟木}和植物的香气，以及橡木带来的香氛。
        \end{CJK} \\ 
        \hline
        \textbf{Llama} & \begin{CJK}{UTF8}{gbsn} 
        这款100\%梅洛特红酒的香气中融合了\textcolor{blue}{百合花}、\textcolor{blue}{草莓}、\textcolor{teal}{樟脑}和\textcolor{teal}{地中海灌木丛}的气息，伴随着橡木驱动的香料。
        \end{CJK}\\
        \hline
        \textbf{NLLB} & \begin{CJK}{UTF8}{gbsn} \textcolor{red}{红虹}，\textcolor{red}{树}，和\textcolor{red}{地中海洗刷}的香气与木驱动的香料混合在这个100\%的\textcolor{red}{梅罗特酒}上。 \end{CJK} \\
        \hline
        \textbf{ChatGPT} & \begin{CJK}{UTF8}{gbsn} 
            这款100\%梅洛红酒散发出\textcolor{teal}{鸢尾花}、\textcolor{teal}{覆盆子}、\textcolor{teal}{樟脑}和\textcolor{teal}{地中海灌木}的香气，并与橡木带来的香料味交织在一起。
        \end{CJK} \\
        \hline
    \end{tabularx}
    \caption{Comparison of Translations from Different Models. Red text shows the wrong translation, teal text shows the correct literal translation, blue text shows the adapted translation and brown text shows unsure adapted translation.}
    \label{tab:detailed_example}
\end{table*}

In Table \ref{tab:detailed_example}, we present a comparative analysis of multiple machine translation models applied to an English-to-Chinese translation example which is shown in Figure \ref{fig:example}. Each row in the table represents the output of a different model, with the original English input provided for reference. To facilitate an in-depth evaluation, we systematically annotate various translation characteristics, including direct translations, adapted phrasings, and potential errors.

To facilitate evaluation, we use color coding to distinguish different translation characteristics: red indicates incorrect translations, teal represents accurate literal translations, blue highlights adapted translations that maintain meaning while improving fluency, and brown marks uncertain adaptations.

This comparison reveals variations in how different models interpret and translate key terms, particularly in handling domain-specific vocabulary such as "Merlot" and "Mediterranean scrub." Some models exhibit direct translation errors, while others apply adaptive strategies to enhance readability. By analyzing these differences, we can better understand the strengths and limitations of current machine translation systems.

\section{Jaccard Similarity Analysis Example}
\label{sec:Jaccard}
‘Fully developed, the nose, with its spice box, tobacco leaf, forest floor, and cigar wrapper calls your name. Medium-bodied, fresh, vibrant, round, and packed with dark currants and red plums, this is ready to go. If you have a bottle, there is no reason to hold this any longer. It will not improve.’

For the above review, we manually extract the Exact Aromas [spices, tobacco, plum, black currant] which can easily be done. And then we use The Wine Aroma Wheel By Aromaster mapping them to [spices, toasted, stone fruit, red berries](Aroma Subfamilies) and [maturation for oak barry, Fruity red Wine]. Here spices and toasted all mapped in [maturation for ‘oak barry and stone fruit’, red berries all mapped in ‘Fruity red Wine’. And based on these three sets we compute Jaccard similarity.

\section{Quantitative analysis}
\label{sec:Case}
Cross-lingual translation of culturally specific concepts is challenging, as it requires balancing linguistic accuracy with cultural adaptation. Many models tend to rely on literal translation, which may not always convey the intended meaning naturally. To examine this, we evaluate each model's literal translation rate for culturally embedded flavor descriptors from the CulturalWR test set. For instance, in English-to-Chinese translation, `thyme' is considered an English-specific concept. We count occurrences of related terms such as `gooseberry', `thyme', and `rosemary' in English wine reviews ($c_{source}$) and record how often they are directly translated in the corresponding Chinese reviews ($c_{target}$) from model predictions. The literal translation rate is then calculated as $\frac{c_{target}}{c_{source}}$. To further assess translation quality, we conduct a bidirectional test on the five most common wine-related terms. This ensures that models not only preserve culturally specific terms when translating in one direction but also effectively map key wine descriptors between languages. Comparing accuracy across models provides insights into their ability to maintain semantic fidelity, crucial for expert-level wine translations.

\begin{figure*}[ht]
    \centering
    \includegraphics[width=0.8\linewidth]{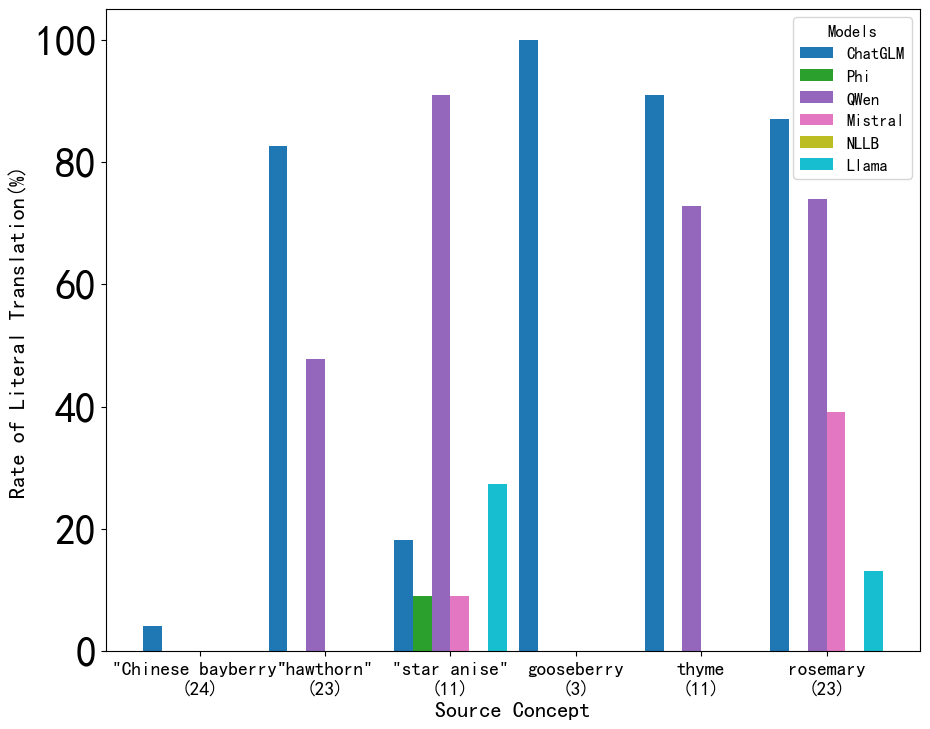}
    \caption{Analysis of the translation of specific concepts by the different models on the test data.}
    \label{fig:Ana_Apatation}
\end{figure*}

As shown in Figure \ref{fig:Ana_Apatation}, we analyze six culturally relevant concepts, three common in Chinese culture and three in Western culture. Results show significant differences in literal translation rates across models ChatGLM and Qwen, with a stronger emphasis on Chinese-language data, exhibit higher literal translation rates, prioritizing structural fidelity over adaptation. For Western cultural concepts, Llama and Mistral show some degree of accurate literal translation, though performance varies. Notably, no model directly translates `waxberry', likely due to its regional specificity and the absence of a widely recognized equivalent in Western languages. NLLB struggles with all six concepts, highlighting potential NMT limitations. Furthermore, while Llama and Mistral do not achieve the highest literal translation rates, they demonstrate a tendency to adapt culturally specific terms rather than translate them directly. For example, they often map `raspberry' to other red berries such as `blueberry' and `blackberry' and replace `thyme' with similar spices like `bay leaves' and `cinnamon'. These adaptations align with the categorization in the Wine Aroma Wheel, as both the substituted berries and spices belong to the same Aroma Subfamilies. 
These findings are consistent with Table \ref{tab:human_eval}, where ChatGLM and Qwen score higher in Faithfulness of Information, while Llama and Mistral perform better in Cultural Proximity.

\begin{figure*}[ht]
    \centering
    \includegraphics[width=0.8\linewidth]{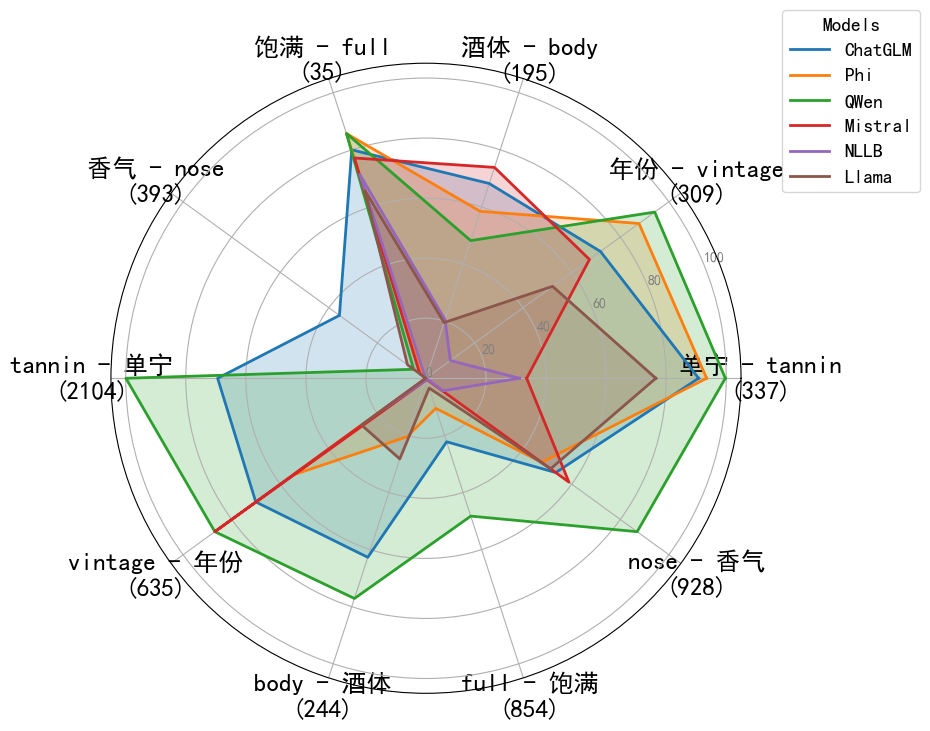}
    \caption{Analysis of the translation of specific terminology by the different models on the test data.}
    \label{fig:Ana_T}
\end{figure*}

We further assess how well models handle wine terminology, which often has industry-specific meanings distinct from general usage. As shown in Figure \ref{fig:Ana_T}, ChatGLM and Qwen perform well in translating these terms, while Phi also ranks highly, even leading for `full'. However, challenges arise with terms such as `nose', which refers to a wine’s aroma in professional contexts. The phrase "on the nose" describes a wine’s bouquet, yet most models translate `nose' directly, failing to capture its specialized meaning. This highlights the challenge of translating wine-specific terms beyond their literal meanings and underscores the importance of integrating domain knowledge into machine translation models.


